\documentclass[lettersize,journal]{IEEEtran}
\usepackage{amsmath,amsfonts}
\usepackage{algorithmic}
\usepackage{algorithm}
\usepackage{array}
\usepackage[caption=false,font=normalsize,labelfont=sf,textfont=sf]{subfig}
\usepackage{textcomp}
\usepackage{stfloats}
\usepackage{url}
\usepackage{verbatim}
\usepackage{graphicx}
\usepackage{cite}

\usepackage{amsmath,graphicx}
\usepackage{adjustbox}
\usepackage{url}
\usepackage{array}
\usepackage{multirow}
\usepackage{amssymb}
\usepackage{pifont}
\usepackage{pgfplots}
\usepackage{multirow}
\usepackage{booktabs}
\usepackage{makecell}
\usepackage{tikz}
\usepackage{tipa}
\usepackage{xcolor}
\usepackage{caption}
\usepackage{subcaption}
\usepackage{lipsum}
\usepackage[table]{xcolor}
\usepackage{arydshln}
\newcommand{\cmark}{\ding{51}}
\newcommand{\xmark}{\ding{55}}
\usepackage{tcolorbox}
\definecolor{lightblue}{rgb}{0.93,0.95,1.0} 

\begin{document}

\title{GCAgent: Long-Video Understanding via Schematic and \\ Narrative Episodic Memory}

\author{Jeong~Hun~Yeo$^{\dagger}$ \quad Sangyun Chung$^{\dagger}$ \quad Sungjune Park \quad Dae Hoe Kim \quad  Jinyoung Moon \quad Yong Man Ro$^{*}$
\thanks{J. H Yeo, S. Chung, S. Park, and Y. M. Ro are with the Integrated Vision and Language Lab., School of Electrical Engineering, Korea Advanced Institute of Science and Technology (KAIST), 291 Daehak-ro, Yuseong-gu, Daejeon, 34141, Republic of Korea (e-mail: sedne246@kaist.ac.kr; jelarum@kaist.ac.kr; sungjune-p@kaist.ac.kr; ymro@kaist.ac.kr). D. H. Kim and J. Moon are with the Visual Intelligence Research Section, Superintelligence Creative Research Laboratory, Electronics and Telecommunications Research Institute (ETRI), Republic of Korea (e-mail: dhkim19@etri.re.kr, jymoon@etri.re.kr). $^{*}$Corresponding author: Y. M. Ro (fax: 82-42-350-5494)  $^{\dagger}$ Both authors contributed equally to this work\protect\\}}



\maketitle

\begin{abstract}
Long-video understanding remains a significant challenge for Multimodal Large Language Models (MLLMs) due to inherent token limitations and the complexity of capturing long-term temporal dependencies. Existing methods often fail to capture the global context and complex event relationships necessary for deep video reasoning. To address this, we introduce GCAgent, a novel Global-Context-Aware Agent framework that achieves comprehensive long-video understanding. Our core innovation is the Schematic and Narrative Episodic Memory. This memory structurally models events and their causal and temporal relations into a concise, organized context, fundamentally resolving the long-term dependency problem. Operating in a multi-stage Perception-Action-Reflection cycle, our GCAgent utilizes a Memory Manager to retrieve relevant episodic context for robust, context-aware inference. Extensive experiments confirm that GCAgent significantly enhances long-video understanding, achieving up to 23.5\% accuracy improvement on the Video-MME Long split over a strong MLLM baseline. Furthermore, our framework establishes state-of-the-art performance among comparable 7B-scale MLLMs, achieving 73.4\% accuracy on the Long split and the highest overall average (71.9\%) on the Video-MME benchmark, validating our agent-based reasoning paradigm and structured memory for cognitively-inspired long-video understanding.
\end{abstract}


\ifCLASSOPTIONcompsoc
\IEEEraisesectionheading{\section{Introduction}\label{sec:introduction}}
\else
\section{Introduction}
\label{sec:introduction}
\fi

With the explosive growth of video-based social media and platforms, video has become the dominant medium in our daily lives, shaping communication, entertainment, and education. As millions of new videos are generated and shared every day, efficient and accurate video processing is now essential for extracting and analyzing meaningful information from this vast and rapidly expanding content. With this background, the rise of Multimodal Large Language Models(MLLMs)~\cite{Maaz2023VideoChatGPTTD, Lin2023VideoLLaVALU, Xu2024PLLaVAP, bai2023qwen, bai2025qwen2, chen2024expanding, Cheng2024VideoLLaMA2A, lin2023video, zhang2023video, xu2025slowfast} has attracted increasing attention. By directly integrating video perception into LLMs, these models provide a promising solution for understanding and interacting with complex video content.

Despite remarkable progress in MLLMs, significant challenges remain in handling long videos, primarily due to the computational burden of modeling long-term temporal context. To address this issue, earlier efforts focus on improving the intrinsic capacity of MLLMs, primarily by extending context length~\cite{team2024gemini, chen2024longvila, wang2024longllava} or reducing token usage through compression of visual embeddings at the feature  level~\cite{fei2024video, li2024llama, shu2025video, song2024moviechat, weng2024longvlm, zeng2024timesuite, zhang2023video}. These approaches aim to process longer video inputs within a single model context window by optimizing the internal efficiency of MLLMs. More recent studies have increasingly shifted toward agent-based paradigms~\cite{wang2024videoagent, chen2025lvagent, yuan2025videodeepresearch, fan2024videoagent, yang2024vca, jeong2025videorag}, which integrate external reasoning mechanisms, retrieval modules, and collaborative planning to handle long videos more effectively. In this paradigm, agents autonomously plan how to retrieve and organize query-related information, alleviating the context length limitation.

Although retrieving query-related information enables more efficient long-video understanding, its capability remains limited. In particular, it falls short compared to the human way of maintaining and organizing global context across the entire video while selectively attending to query-relevant information, leaving clear room for improvement. Recent works in video understanding have also highlighted the role of temporal structural modeling and long-term action reasoning~\cite{wang2025multimodal, xu2025quality, qian2023locate, fang2023hierarchical}. Cognitive psychology and cognitive science \cite{bartlett1995remembering, zacks2007event} suggest that humans comprehend and remember events by constructing schematic and narrative structures. Here, schematic structure refers to abstract event templates (e.g., roles, typical situation frames), whereas narrative structure denotes temporally and causally ordered event sequences. These structures enable humans to more efficiently integrate new information and perform downstream tasks. Extending this insight to long video understanding, if MLLMs could build and leverage a comprehensive understanding of the global context while using query-related information to generate answers, they would likely demonstrate enhanced task performance in a more human-like manner. Bridging this cognitive insight with computational models offers a clear path toward more human-like video understanding. 

In this paper, we introduce GCAgent, a global-context-aware agent framework for long-video understanding. Specifically, GCAgent grounds global context representation in schematic and narrative structures, mirroring how humans construct and maintain situation-level understanding. At the same time, it preserves the strengths of conventional agent-based methods in retrieving query-related information. By combining these advantages, our framework significantly enhances the ability of MLLMs to understand and reason over long videos in query-driven interaction scenarios. To realize this awareness, GCAgent is composed of two complementary agents: (i) Memory Manager Agent (LLM-based), which constructs and maintains the global context before any query arrives. To this end, it primarily utilizes speech transcripts as input. The agent first detects event boundaries and segments the transcript into event-level units. Each event unit is then abstracted to extract roles and situation-level patterns, yielding schematic structures. Finally, temporal and causal relationships across these discrete events are inferred to build narrative structures, forming the episodic memory of the video. Concretely, each event-level unit becomes an episode entry in the episodic memory. When speech is unavailable or insufficient, visual captions can be optionally incorporated as supplementary evidence. (ii) Reasoning Agent (MLLM-based), which leverages both the constructed narrative structures and the query-related multimodal information to perform context-aware reasoning and answer the user query.

Once the episodic memory is structured, the agents collaborate through a three-stage paradigm to address user queries: (i) Perception, which locates query-relevant segments; (ii) Action, which performs reasoning conditioned on the global context; and (iii) Reflection, which updates memory based on the reasoning outcome. Upon query arrival, the memory manager agent switches from global context construction to query-conditioned retrieval. In the Perception stage, the memory manager agent first retrieves key spans in the speech transcripts that are most relevant to the query, maps them to temporal boundaries to extract the corresponding video segments, and forwards the clipped segments together with the transcript excerpts to the next stage. During the Action stage, the reasoning agent grounds its inference in the global schematic and narrative context maintained in episodic memory, while generating answers based on the paired transcript spans and their corresponding video segments retrieved in the previous stage. Finally, in the Reflection stage, the memory manager agent updates episodic memory by integrating the reasoning outcome and the textual descriptions generated during Action, thereby enriching the stored narrative context for subsequent queries.

The contributions of this paper can be summarized as:
\begin{itemize}
    \item \textbf{Global-Context-Aware Agent Framework}
    We propose GCAgent, the first agent-based framework for long-video understanding that constructs global context as episodic memory before queries. This closes the gap between global context modeling and query-conditioned retrieval.
    \item \textbf{Schematic and Narrative-Structured Memory Representation}
    We instantiate episodic memory with schematic and narrative structures, enabling explicit global context organization rather than shallow snippet-level summaries.
    \item \textbf{Effectiveness and Efficiency of Episodic Memory:}
    Our episodic-memory design yields substantial performance gains: GCAgent improves accuracy by up to 23.5\% higher accuracy on VideoMME Long split, over a vanilla Qwen2.5-VL baseline (i.e., same backbone MLLM without our agent).
\end{itemize}

\section{Related Works}
\subsection{Long Video Understanding with MLLMs}
\subsubsection{Multimodal Large Language Models} Recent progress of MLLMs in video domain has been accelerated by several key accomplishments: First, advancements in visual encoders \cite{Radford2021LearningTV, Zhai2023SigmoidLF, Dehghani2023PatchNP, Beyer2022FlexiViTOM} have substantially improved video representation quality. In particular, CLIP-based Vision Transformer (ViT)~\cite{Radford2021LearningTV} encoders have become the dominant backbone for MLLMs, as their strong cross-modal alignment and scalable architecture enable effective integration of video features into LLMs. More recently, several ViT variants~\cite{Radford2021LearningTV, liu2024oryx} extend this line of work by supporting arbitrary input resolutions and improving computational efficiency, thereby further enhancing the adaptability of visual backbones to diverse video sources. Second, the release of open-source LLMs \cite{bai2023qwen, touvron2023llama, chiang2023vicuna, jiang2024mixtral} has democratized model development and spurred rapid innovation in multimodal learning, making it feasible to adapt general-purpose LLMs for video-centric tasks. Third, the creation of large-scale video instruction-following datasets \cite{Zhang2024DirectPO, Chen2024ShareGPT4VideoIV, Zhang2024VideoIT, lu2023show} has enabled supervised alignment between video content and language, facilitating the emergence of instruction-following capabilities in MLLMs.

\subsubsection{Long Video Understanding}
With the progress of MLLMs, notable improvements have been achieved in video understanding tasks such as question answering, captioning, and summarization, demonstrating the strong potential of multimodal reasoning over video content. Building on these successes, research attention has recently shifted toward the more challenging problem of long video understanding, accompanied by the introduction of benchmarks designed to evaluate and verify MLLM's capabilities \cite{Wu2024LongVideoBenchAB, fu2024video}. However, this direction remains difficult due to the limited context windows of LLMs \cite{Xue2024LongVILASL} and the large number of visual tokens required for video representation. To address these limitations within the model itself, two major lines of research have been explored, both aiming to enhance the intrinsic capacity of MLLMs without relying on external tools.
The first extends the context length of LLMs to accommodate longer sequences~\cite{team2024gemini, chen2024longvila, wang2024longllava}. The second reduces the number of visual tokens through compression strategies~\cite{fei2024video, li2024llama, shu2025video, weng2024longvlm, zeng2024timesuite, zhang2023video}, which often exploit redundancies between adjacent frames or compress content at the event level to preserve essential semantics. In parallel, another line of research has sought to improve efficiency at the input level, primarily through heuristic frame sampling or key-frame retrieval strategies~\cite{Wang2019VaTeXAL, Cheng2024VideoLLaMA2A, Liang2024KeyVideoLLMTL, kim2024salova}, which aim to reduce temporal redundancy and enhance processing efficiency before the modeling stage.

Despite these advancements focusing primarily on extending the intrinsic capacity of MLLMs, our work takes a complementary perspective by enhancing long video understanding through external memory organization and agent-based reasoning, which jointly enable structured, narrative-aware comprehension of long videos.

\subsection{Agent-based Long Video Understanding}
Agentic AI~\cite{durante2024agent} aims to construct autonomous systems minimizing human intervention by decomposing complex queries into sub-tasks, making decisions, performing goal-directed planning, and employing external tools to execute actions. In this context, agent-based methods~\cite{wang2024videoagent, chen2025lvagent, yuan2025videodeepresearch, fan2024videoagent, yang2024vca, jeong2025videorag} for long video understanding follow a similar paradigm: upon receiving a query, the agent determines what information is required, formulates a plan to retrieve it, and sequentially invokes external tools to gather and verify relevant evidence. Through this process, the agent achieves strong performance while effectively alleviating the limited context window problem of LLMs.

Concretely, agent-based methods for long video understanding differ in how they design algorithms to identify and refine query-related information within this framework. A first line of research centers on query–information matching, wherein models explicitly align textual queries with visual representations. In particular, CLIP~\cite{Radford2021LearningTV} has been widely adopted to localize the video regions most relevant to the given query~\cite{wang2024videoagent, xie2023openagents, zhao2024longagent}, thereby enabling precise evidence grounding. A complementary direction emphasizes retrieval strategies. Retrieval-Augmented Generation (RAG)~\cite{jeong2025videorag} exemplifies this paradigm by incorporating external tools such as memory banks~\cite{fan2024videoagent, he2024ma} and search engines~\cite{Chen2024ShareGPT4VideoIV, li2024searchlvlms}, which substantially enhance the comprehensiveness of information acquisition. More recently, methods have advanced towards multi-round collaborative pipelines driven by MLLM agents~\cite{chen2025lvagent}, where multiple agents iteratively exchange information and employ majority voting or confidence-based mechanisms to enhance the reliability of final outputs.

\begin{figure*}[t]
\centering
\centerline{\includegraphics[width=17cm]{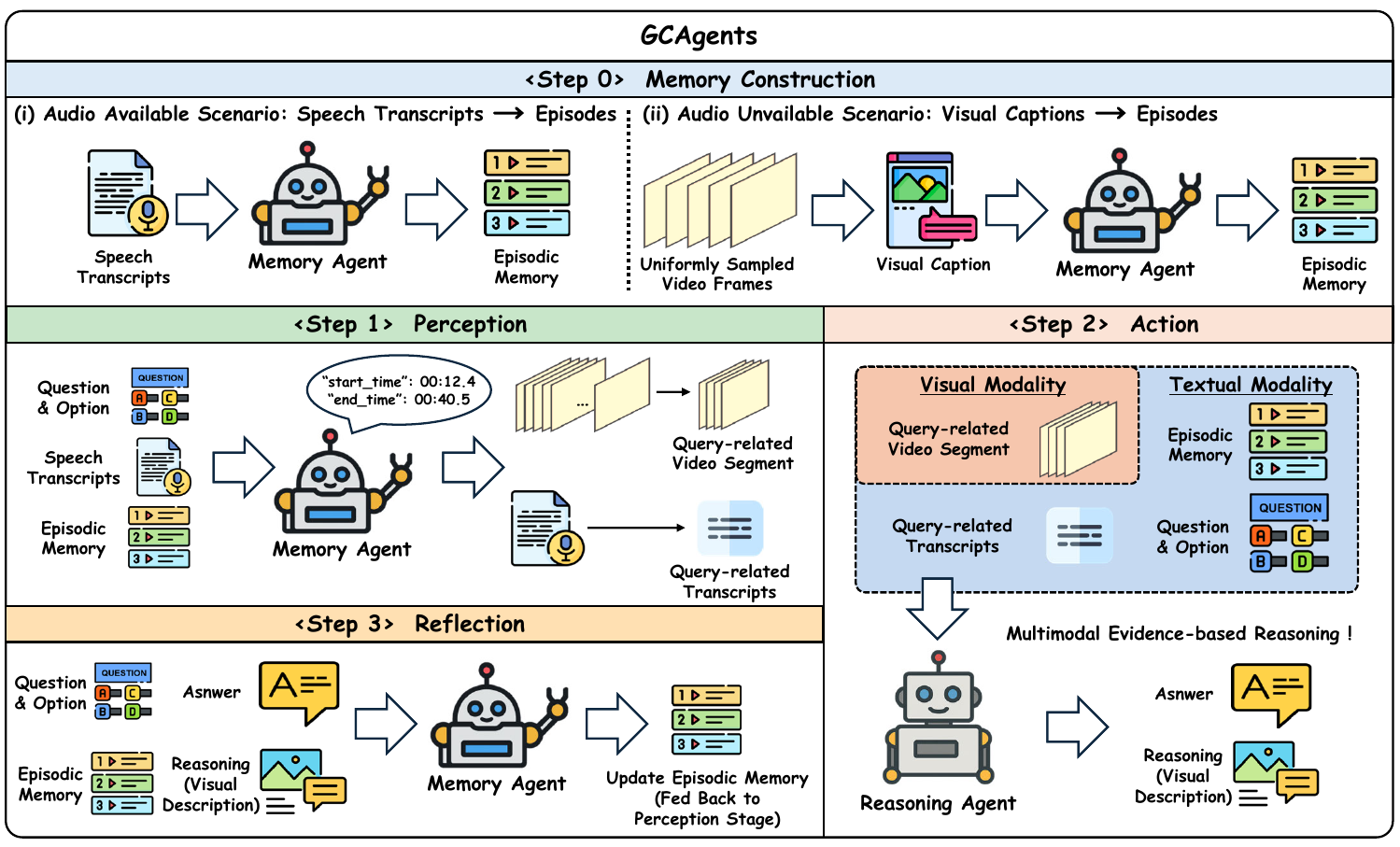}}
\caption{Overview of our GCAgent framework. This framework comprises a memory manager agent handling memory construction and retrieval, and a reasoning agent performing context-aware inference. The process follows four steps: Memory Construction, Perception, Action, and Reflection, to integrate global context with query-related evidence.}
\label{fig:1}
\vspace{-0.4cm}
\end{figure*}

However, these methods primarily focus on retrieving query-related information while paying limited attention to constructing a coherent global understanding of the video. In contrast, our work aims to bridge this gap by enabling agents to organize and utilize episodic memory that captures both schematic and narrative contexts for comprehensive long video understanding.

\section{Method}
We now describe our method, which equips the model with explicit episodic memory before queries and leverages it during query-time reasoning. We first provide an overview of the framework (Sec.~\ref{sec3.1}). We then formalize the episodic memory representation grounded in schematic and narrative structures and describe how the memory manager agent constructs it from speech transcripts (Sec.~\ref{sec3.2}). Finally, we present the full pipeline (Sec.~\ref{sec3.3}), which operates in three phases: Perception, Action, and Reflection.

\subsection{Framework Overview}
\label{sec3.1}
We propose GCAgent, a global-context-aware agent framework for long-video understanding. As illustrated in Fig.~\ref{fig:1}, it consists of two complementary agents. The memory manager agent constructs structured episodic memory and, upon query arrival, retrieves the transcript spans most relevant to the query and maps them to their corresponding video segments. The reasoning agent conditions its inference on this episodic memory, verifies retrieved evidence, and derives answers through logical reasoning. In what follows, we detail our framework in two phases: (i) memory construction performed prior to user queries, and (ii) query-driven reasoning, which follows a perception–action–reflection cycle.

\begin{figure*}[t]
\centering
\centerline{\includegraphics[width=16.5cm]{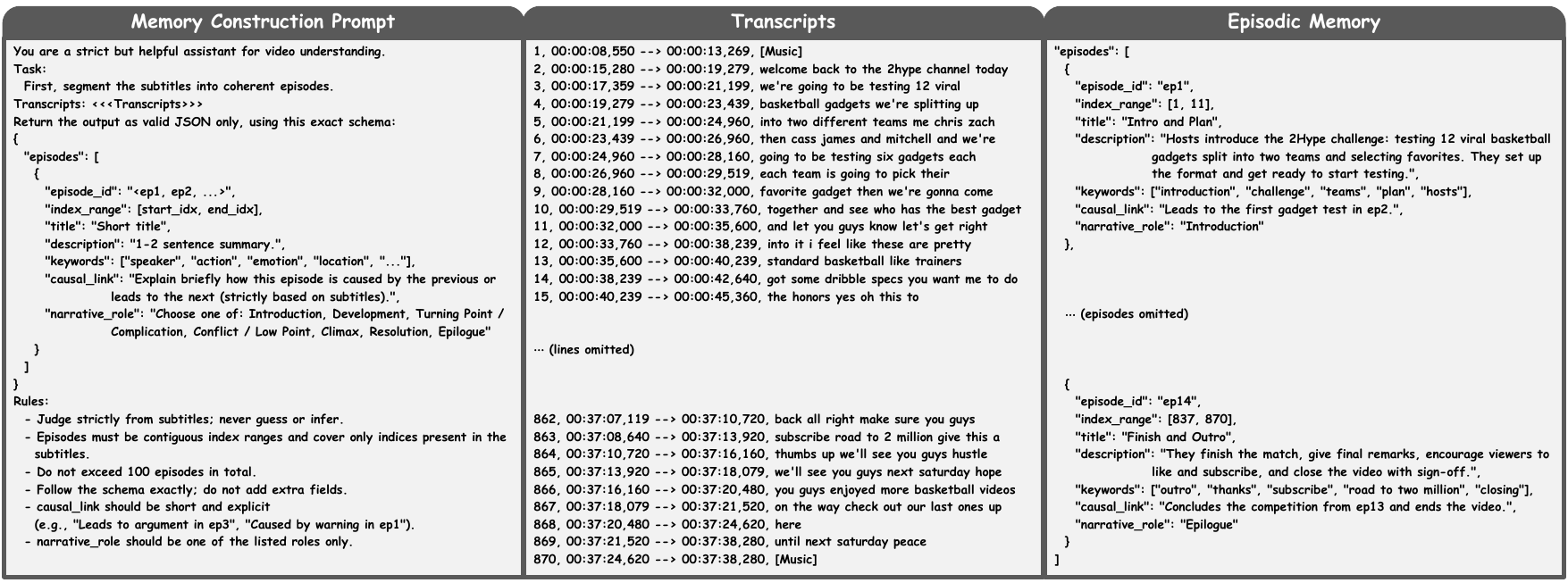}}
\caption{
Example of the episodic memory construction process. The figure illustrates how the Memory Construction Prompt interacts with the speech transcript to generate structured episodic memory representations.
}
\label{fig:2}
\vspace{-0.4cm}
\end{figure*}

\subsection{Memory Construction}
\label{sec3.2}
To organize schematic and narrative structures, we utilize speech transcripts as inputs to the memory manager agent. Specifically, our framework prioritizes transcripts obtained as audio-based subtitles from raw audio. This design choice is motivated by prior studies showing that speech transcripts can be represented with significantly fewer text tokens than video inputs, since the token count of video features grows rapidly with both spatial resolution and temporal duration~\cite{shao2025tokens}.

\subsubsection{Episodic Memory based on speech transcripts} 
Speech transcripts are obtained from raw audio through two audio-based technologies. First, a Voice Activity Detection (VAD)~\cite{sohn1999statistical} module segments the continuous audio stream into distinct speech intervals. Then, an Automatic Speech Recognition (ASR)~\cite{radford2023robust} system generates time-aligned transcripts for each interval, providing accurate textual representations of the spoken content. In our framework, these transcripts serve as the foundation for subsequent memory construction.

The agent performs two key functions during the episodic memory construction process. First, it forms schematic structures by cleaning and abstracting the speech transcripts. Specifically, spoken language naturally contains repetitions, fillers, and discourse markers, making it inherently more compressible than visual streams. We therefore first detect event boundaries to partition the stream into coherent event-level units (i.e., topic shifts), as shown in Fig.~\ref{fig:2}. Each event unit is then abstracted to distill its situation-level meaning. Second, the agent constructs narrative structures across events. In this process, the focus is no longer on summarizing within an event, but on reasoning between events. The agent infers the narrative role of each event-level unit (e.g., introduction, conflict, resolution) based on surrounding discourse and temporal flow. It also identifies causal dependencies across units to organize them into a coherent storyline. Once these roles and relationships are assigned, each event-level unit is finalized as an episode entry in the episodic memory.

\subsubsection{Audio-Unavailable Scenario}
We design memory construction to operate under two modality conditions: (i) audio-available and (ii) audio-unavailable. The previous subsection described the audio-available path. Here, we describe the complementary visual-only path, which applies when videos inherently lack audio (e.g., sports broadcasts without commentary, silent surveillance footage). In this case, the memory construction module switches to a visual path, sampling video frames and generating textual descriptions via an image captioning~\cite{hossain2019comprehensive}. These captions serve as the textual backbone of memory construction under the audio-unavailable condition.

\subsection{Perception–Action–Reflection Pipeline}
\label{sec3.3}
This subsection details the GCAgent’s query-driven reasoning procedure, which unfolds as a Perception–Action–Reflection pipeline. Before detailing each process, we first define the notations used throughout this subsection. We denote the input query as $Q$ and the set of answer options as $O=\{o_1,\dots,o_K\}$. The speech transcripts are written as $T=\{t_1,\dots,t_N\}$, and the episodic memory as $M=\{m_1,\dots,m_L\}$, where each $m_i$ represents a structured episode that encodes both schematic and narrative information.

\subsubsection{Perception}
Recent studies~\cite{chen2025lvagent} demonstrate that retrieving query-related video segments during the perception stage can effectively reduce redundant information interference. Building on this finding, we incorporate such retrieval-based advantages into our framework by enabling the agent to selectively attend to the most visually informative segments. To this end, the memory manager agent switches from global context construction to query-conditioned retrieval. Concretely, it first locates query-relevant spans in the speech transcripts and extracts their time boundaries; these temporal indices are then used to obtain both the corresponding video segments and their matched textual evidence. Formally, the perception step can be written as:
\begin{equation}
    V^{*}, \; T^{*} = \mathcal{A}_{\text{perc}}(Q, O, T, M),
\end{equation}
where $\mathcal{A}_{\text{perc}}(\cdot)$ is the perception module of the memory manager agent. Given $Q$, $O$, $T$, and $M$, it returns the query-relevant video segments $V^{*}$ together with their aligned speech transcripts $T^{*}$. This perception output serves as the input to the subsequent Action and Reflection stages.

\subsubsection{Action}
The goal of the action stage is to answer the query. At this point, the episodic memory $M$ provides global context, while the perception stage has already prepared the query-relevant video segments $V^{*}$ and their textual counterparts $T^{*}$. The reasoning agent then jointly reasons over this multimodal evidence to evaluate how each answer option $O$ aligns with the retrieved context. Formally, the action process can be represented as:
\begin{equation}
    a^{*}, E^{*} = \mathcal{A}_{\text{act}}(Q, O, V^{*}, T^{*}, M),
\end{equation}
where $\mathcal{A}_{\text{act}}(\cdot)$ denotes the action function of the reasoning agent. Here, $a^{*}$ is the predicted answer, and $E^{*}$ denotes the visual explanation (i.e., evidence) that supports this answer. This evidence is then passed to the subsequent stage, where it is used to update the episodic memory.

\subsubsection{Reflection}
The reflection stage focuses on updating the episodic memory based on the visual explanation $E^{*}$ produced during the action stage. At this point, the memory manager agent takes the predicted answer $a^{*}$, the corresponding evidence $E^{*}$, the input query $Q$, the answer options $O$, and the current episodic memory $M$ as inputs. It integrates concise visual summaries back into memory, thereby improving the model’s ability to retain and contextualize visual observations for future queries. Formally, the reflection process can be expressed as:

\begin{equation}
    M' = \mathcal{A}_{\text{refl}}(Q, O, a^{*}, E^{*}, M),
\end{equation}
where $\mathcal{A}_{\text{refl}}(\cdot)$ denotes the reflection function of the memory manager agent, and $M'$ represents the updated episodic memory.

\section{Experimental Setup}
\subsection{Benchmark}
For the experiments, we utilize two video understanding benchmarks. To evaluate the proposed method on long video understanding, we adopt Video-MME~\cite{fu2024video} and LongVideoBench ~\cite{Wu2024LongVideoBenchAB}, which contain long-duration videos of up to one hour and are specifically designed to assess models’ capability in long-term temporal reasoning. In the following subsections, we describe the details of the datasets and implementation details.

\subsubsection{Video-MME}
Video-MME~\cite{fu2024video} consists of 900 manually curated videos totaling 254 hours, accompanied by 2,700 human-annotated question–answer pairs (three questions per video). The dataset is evenly divided into 300 short, 300 medium, and 300 long videos, covering a wide temporal range from 11 seconds to 1 hour (average 1,024 seconds). It spans six major categories: Life Record, Knowledge, Sports, Competition, Film and Television, and Artistic Performance, with a small proportion of multilingual content. Each video includes multi-modal inputs such as frames, subtitles, and audio, allowing evaluation of MLLMs across diverse modalities.

\subsubsection{LongVideoBench}
LongVideoBench~\cite{Wu2024LongVideoBenchAB} is an extreme long video understanding benchmark designed to evaluate models’ capability in handling extended temporal contexts. It contains 500 videos with an average duration of 4,101 seconds, accompanied by 1,549 question–answer pairs. The videos are categorized into six domains, including TV shows, cartoons, documentaries, sports, lifestyle, and event recordings. The benchmark covers five types of reasoning tasks, namely entity recognition, event understanding, key information retrieval, temporal grounding, and summarization, providing a comprehensive evaluation setting for long video understanding.

\subsection{Implementation details}
Our framework consists of two main components: a memory manager agent and a reasoning agent. The memory manager agent is implemented using the GPT-5.1 Mini model, while the reasoning agent employs the Qwen2.5-VL 7B model. The maximum input length for the Qwen2.5-VL 7B model is set to 32 frames. All answer generation processes are performed through the reasoning agent. For the experiments, we utilize eight NVIDIA A6000 GPUs (48GB each). Speech transcripts are obtained from the subtitles provided by each dataset, which are originally generated via YouTube’s automatic captioning system.

\begin{table*}[t]
\centering
\caption{Comparison with recent state-of-the-art methods on LVBench and Video-MME. Gray-shaded entries denote models utilizing MLLMs larger than 70B, excluded from direct comparison with 7B-scale agentic frameworks. All results are reported in accuracy (\%).}
\renewcommand{\arraystretch}{0.9}
\renewcommand{\tabcolsep}{2.0mm}
\resizebox{0.9\linewidth}{!}{
\begin{tabular}{lcccccc}
\toprule
\multirow{2}{*}{\textbf{Model}} &
\multirow{2}{*}{\textbf{VLM Params}} &
\multirow{2}{*}{\textbf{LongVideoBench}} &
\multicolumn{4}{c}{\textbf{Video-MME}} \\
\cmidrule(lr){4-7}
 & & & \textbf{Short} & \textbf{Medium} & \textbf{Long} & \textbf{Average} \\
\midrule
\rowcolor{gray!15} \multicolumn{7}{l}{\textit{Proprietary MLLMs}} \\
GPT-4o~\cite{hurst2024gpt} & - & 66.7 & 82.8 & 76.6 & 72.1 & 77.2 \\
Gemini-1.5-Pro~\cite{team2024gemini} & - & 64.0 & 84.5 & 81.0 & 77.4 & 81.3 \\
\midrule
\rowcolor{gray!15} \multicolumn{7}{l}{\textit{Direct Reasoning MLLMs}} \\
ShareGPT4Video-8B~\cite{Chen2024ShareGPT4VideoIV} & 8B & 39.7 & 53.6 & 39.3 & 37.9 & 43.6 \\
VideoChat2-7B~\cite{li2024mvbench} & 7B & 39.3 & 52.8 & 39.4 & 39.2 & 43.8 \\
InternVL-2.5-8B~\cite{chen2024expanding} & 8B & 60.0 & - & - & - & 66.9 \\
Qwen2-VL-7B~\cite{Wang2024Qwen2VLEV} & 7B & 55.6 & - & - & - & 69.0 \\
Qwen2.5-VL-7B~\cite{bai2025qwen2} & 7B & 56.0 & - & - & - & 71.6 \\
LongVA-7B~\cite{zhang2024long} & 7B & 51.3 & 61.6 & 53.6 & 47.6 & 54.3 \\
LongVU-7B~\cite{shen2024longvu} & 7B & - & - & - & 59.5 & 60.6 \\
LLaVA-Video-7B~\cite{zhang2024video} & 7B & 58.2 & - & - & - & 69.7 \\
\midrule
\rowcolor{gray!15} \multicolumn{7}{l}{\textit{Agentic Frameworks}} \\
VideoTree~\cite{wang2025videotree} & 7B & - & 67.8 & 59.9 & 54.2 & 60.6 \\
VideoRAG~\cite{luo2024video} & 7B & 58.4 & 66.4 & 60.2 & 59.8 & 62.1 \\
DrVideo~\cite{ma2025drvideo} & 7B & - & - & - & 71.7 & - \\
\textcolor{gray!60}{VideoRAG~\cite{luo2024video}} & \textcolor{gray!60}{72B} & \textcolor{gray!60}{65.4} & \textcolor{gray!60}{81.1} & \textcolor{gray!60}{72.9} & \textcolor{gray!60}{73.1} & \textcolor{gray!60}{75.7} \\
\textcolor{gray!60}{LVAgent~\cite{chen2025lvagent}} & \textcolor{gray!60}{78B} & \textcolor{gray!60}{-} & \textcolor{gray!60}{90.7} & \textcolor{gray!60}{87.6} & \textcolor{gray!60}{81.7} & \textcolor{gray!60}{86.6} \\
\midrule
Ours  & 7B & 55.0 & 72.6 & 69.8 & 73.4 & 71.9 \\
\bottomrule
\end{tabular}
}
\label{tab:1}
\end{table*}

\section{Experimental Results}
\subsection{Comparison with the state-of-the-art methods}
To assess the overall effectiveness of our framework, we conduct a comprehensive comparison against recent state-of-the-art methods on the LongVideoBench and Video-MME benchmarks. We group the compared methods into three categories: Proprietary MLLMs, Direct-Reasoning MLLMs, and Agentic Frameworks. The Proprietary group contains closed-source state-of-the-art models and provides upper-bound references. The Direct-Reasoning group evaluates the native visual reasoning ability of MLLMs without any auxiliary modules (we adopt numbers from the original papers). For the Agentic group, we compare VideoTree, VideoRAG, and DrVideo, all of which adopt 7B-based MLLMs as their core backbone.

In Table 1, our proposed method achieves 55.0\% on LongVideoBench and 72.6\% / 69.8\% / 73.4\% / 71.9\% for the short, medium, long, and average settings on Video-MME, respectively. When compared with the first group (Proprietary MLLMs), models such as GPT-4o and Gemini-1.5-Pro achieve average accuracies above 70\% on both benchmarks, which indicates overall superior performance. However, on the long split of the Video-MME benchmark, our method, despite being a 7B-scale model, achieves 73.4\%, demonstrating competitive performance comparable to GPT-4o (72.1\%) and Gemini-1.5-Pro (77.4\%). Although its overall average accuracy is lower than that of larger proprietary models, it shows strong efficiency in long-term video understanding even at a smaller scale. In comparison with the Direct Reasoning MLLMs group, our method shows clear advantages. In particular, on Video-MME it achieves the highest average accuracy (71.9\%) among all 7B-scale models, demonstrating strong multimodal reasoning over extended temporal contexts. Finally, compared to the Agentic Frameworks group, our method shows clear gains on Video-MME. In the 7B-scale setting, it consistently surpasses VideoTree, VideoAgent, and VideoRAG across all three splits. In particular, on the long split , our method reaches 73.4\%, whereas existing 7B-based agentic systems remain below ~60\% (VideoTree) or stay in the low-60\% range (VideoRAG). This demonstrates that our approach provides markedly stronger long-context reasoning, showcasing the effectiveness of global-context awareness, especially in long-video scenarios.

\subsection{Ablation Study}
\subsubsection{Effect of Evidence Selection and Memory Structure in Long-Video Reasoning}
To quantify how different information sources affect long-video reasoning, we conduct an ablation study on the Video-MME (Long split) benchmark using Qwen2.5-VL-7B. We organize the analysis along three modality conditions: (1) vision-only, (2) text-only, and (3) vision + text. Within each modality condition, we evaluate different input evidence compositions. Here, input evidence composition refers to which specific visual or textual sources are provided to the reasoning agent, allowing us to isolate which evidence contributes most to reasoning. Throughout all ablation variants, the reasoning agent is instantiated as Qwen2.5-VL-7B. All ablation results are summarized in Table~\ref{tab:2}.

When the reasoning agent is only allowed to observe the video (vision-only), the baseline Qwen2.5-VL using 32 uniformly sampled frames achieves 49.9\%. 
Replacing this input with the query-relevant video segment yields 56.1\%, a +6.2 point gain. This observation indicates that the memory manager agent effectively identifies the query-relevant time span, from which the corresponding video segment is extracted.

\begin{table}[t]
\caption{Ablation study on the effect of evidence composition and memory design, measured on Video-MME (Long split). QR: query-related. Narrative structure indicates causal links and narrative roles. Unless explicitly noted, the episodic memory refers to only the schematic structure.}
\centering
\renewcommand{\arraystretch}{0.95}
\renewcommand{\tabcolsep}{4.5mm}
\resizebox{0.999\linewidth}{!}{
\begin{tabular}{lc}
\toprule
\textbf{Configuration} & \textbf{Accuracy (\%)} \\
\midrule
\rowcolor{gray!15} \textbf{Vision Only} & \\
\quad Uniform Sampling (Baseline) & 49.9 \\
\quad QR Video Segment & 56.1 \\
\midrule
\rowcolor{gray!15} \textbf{Text Only} & \\
\quad Full Speech Transcript & 65.0 \\
\quad QR Speech Transcript & 65.3 \\
\quad Full Speech Transcript + Episodic Memory & 64.7 \\
\quad QR Speech Transcript + Episodic Memory & 67.0 \\
\midrule
\rowcolor{gray!15} \textbf{Vision + Text (based on QR Video Segment)} & \\
\quad + QR Speech Transcript & 68.8 \\
\quad + Memory & 64.6 \\
\quad + QR Speech Transcript + Episodic Memory & 71.0 \\
\quad + QR Speech Transcript + Episodic Memory (w/ narrative structure) & 73.4 \\
\bottomrule
\end{tabular}
}
\label{tab:2}
\vspace{-0.4cm}
\end{table}

When the reasoning agent is only allowed to read text (text-only), using the entire speech transcript yields 65.0\%, while restricting the input to the query-related speech transcript gives a slightly higher 65.3\%. Introducing episodic memory further clarifies the effect of context: combining memory with the full speech transcript decreases performance to 64.7\%, but combining memory with the query-related speech transcript improves it to 67.0\%. This behavior is natural given that the episodic memory is distilled from the full speech transcript. When both are provided together, the agent effectively receives the same global information twice, which adds redundancy rather than benefit. In contrast, pairing episodic memory with the query-related speech transcript is complementary: the speech transcript provides the local evidence required for answering, while the memory supplies compact global context.

When both modalities are used together (vision + text), we restrict the visual input to the query-relevant video segment. Adding the corresponding query-related speech transcript further boosts accuracy to 68.8\%, showing that text provides complementary evidence that grounds what is visually observed. Injecting episodic memory improves performance again to 71.0\%, and introducing narrative structure yields the final 73.4\%. This progression suggests that once the query-related local evidence (video + speech transcript) is established, the understanding of global context becomes useful.

\begin{table*}[t]
\caption{Evaluation of category-wise Performance, measured on the Video-MME (Long split). Answering module: Qwen2.5-VL-7B. Inference: 32 video frames. QR: Question-related. Gray highlight indicates the “Multilingual” category, which exhibits a distinct performance trend compared to others. The Memory configuration includes the proposed narrative structure (w/ narrative structure).}
\centering
\renewcommand{\tabcolsep}{1.6mm}
\begin{tabular}{lccccccc}
\toprule
\textbf{Configuration} & \textbf{Knowledge} & \textbf{Film \& TV} & \textbf{Sports} & \textbf{Artistic Life} & \textbf{Multilingual} & \textbf{Competition / Record} & \textbf{Overall} \\
\midrule
QR Video Segment & 60.4 & 52.5 & 61.3 & 50.8 & \cellcolor{gray!15}56.7 & 54.3 & 56.7 \\
+ QR Transcript & 76.3 (+15.9) & 62.5 (+10.0) & 71.3 (+10.0) & 62.5 (+11.7) & \cellcolor{gray!15}70.0 (+13.3) & 61.4 (+7.1) & 68.1 (+11.4) \\
+ Memory & 74.4 (+14.0) & 60.8 (+8.3) & 68.7 (+7.4) & 65.8 (+15.0) & \cellcolor{gray!15}50.0 (-6.7) & 63.8 (+9.5) & 67.2 (+10.5) \\
+ QR Transcript, Memory & 81.5 (+21.1) & 66.7 (+15.2) & 74.0 (+12.7) & 70.8 (+20.0) & \cellcolor{gray!15}60.0 (+3.3) & 70.0 (+15.7) & 73.4 (+16.7) \\
\bottomrule
\end{tabular}
\label{tab:4}
\end{table*}

\subsection{Analysis}
\subsubsection{Analyzing Category-Wise Performance}
To further analyze the performance of our method across different domains, we conduct experiments on the Video-MME (Long split) benchmark. In particular, we report results on six categories: Knowledge, Film \& TV, Sports, Artistic Life, Multilingual, and Competition / Record, as summarized in Table~\ref{tab:4}.

We first establish a baseline by evaluating the reasoning agent using only the query-related video segment for each category, achieving an average accuracy of 56.7\%. Subsequently, we evaluate performance by adding the query-related transcript and episodic memory separately, as summarized in the second and third rows of Table~\ref{tab:4}. When incorporating the query-related transcript alone, the model achieves an average improvement of 11.4\% across the six categories, while adding only memory results in a 10.5\% gain on average. The result shows that incorporating either query-related transcripts or episodic memory consistently improves performance across categories. Furthermore, combining the query-related transcript and episodic memory produces a synergistic effect in most categories. For instance, the Knowledge category achieves a remarkable 21.1\% improvement, exceeding the gains obtained when each component is used individually.

Notably, the Multilingual category shows a different trend. When episodic memory is applied, performance degrades, suggesting that it may interfere with multilingual reasoning. The following subsection conducts a qualitative analysis of the question–answer pairs within this category to identify the underlying factors responsible for the performance drop.

\begin{figure}[t]
\centering
\centerline{\includegraphics[width=8cm]{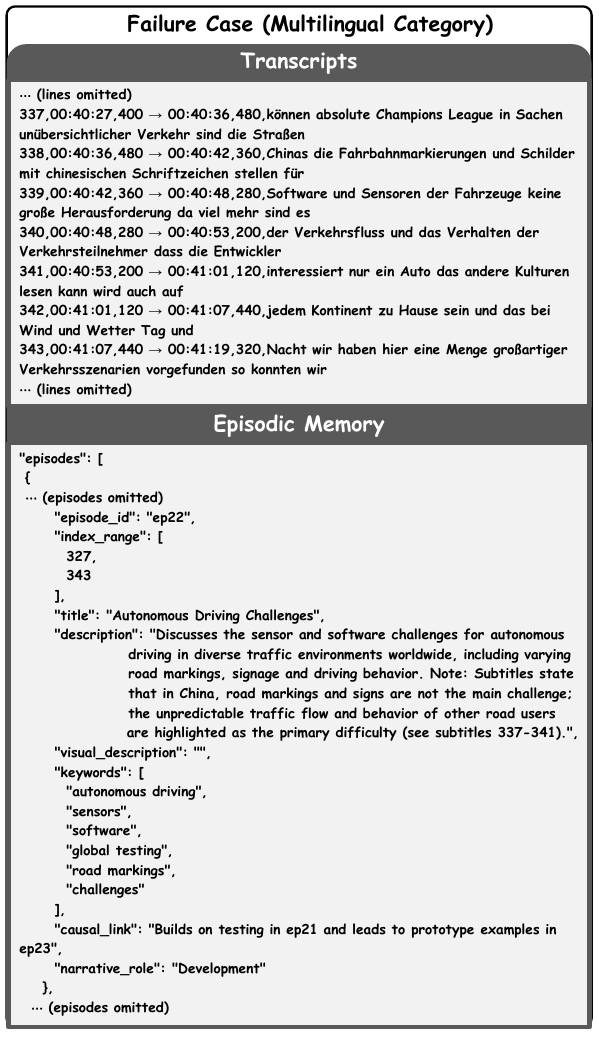}}
\caption{Illustration of multilingual performance degradation. The figure presents the transcript segment (lines 337–341) and its corresponding memory summary.}
\label{fig:3}
\end{figure}

\subsubsection{Detailed Analysis of Multilingual Performance Degradation}

To better understand the degradation pattern observed in the Multilingual category, we analyze one illustrative case where memory addition leads the model to an incorrect answer. The question is: \textit{"In line with the video evidence, what is the primary challenge for developers of autonomous driving technology, as highlighted by the testing conducted in China?"} Transcriptions from the related video segments (lines 337-341) and the corresponding episodic memory summary are shown in Figure 3. In the transcripts, the narrator clarifies that road markings and signage are not the main issue in China; instead, the unpredictable traffic flow and behavior of other road users constitute the primary difficulty, corresponding to option B. However, the episodic memory summary generalizes the scene as “discussing sensor and software challenges in diverse traffic environments worldwide,” thereby flattening the original contrast (“A is not the problem; B is”) into a neutral enumeration (“A, B, and C are challenges”). Consequently, the model incorrectly selects option A instead of B.

Similar to this example, we observe that in the Multilingual category, segments containing non-English speech, such as German in this case, though not necessarily limited to it, may be automatically translated into English during the memory summarization process. This translation can potentially introduce semantic distortion or subtle loss of contextual nuance, which may alter the intended meaning of the original utterance. In contrast, such issues are not observed in the other five categories, which all consist of English content where both the transcript and the memory summary appear to remain consistent in language and meaning, and no notable performance degradation is observed. These observations suggest that multilingual content may introduce additional complexity to the memory construction process, which could contribute to the observed performance degradation.

\subsubsection{Analyzing Token Compression Efficiency of Memory}
To further examine the efficiency of the proposed memory mechanism, we analyze the token usage of the full speech transcripts and the constructed memory summaries on the Video-MME benchmark. As demonstrated in Sec.~5.2.1, the memory can replace full speech transcript while achieving nearly equivalent performance. Here, we quantitatively evaluate its efficiency in providing global contextual information by measuring the ratio between full transcript tokens and memory tokens across different video duration ranges. The results are presented in Table~\ref{tab:5}. Following the split scheme of Video-MME (Short, Medium, and Long), 
we group videos into three duration ranges: 0–2 minutes, 4–15 minutes, and 30–60 minutes. In the case of short videos, the full transcripts are very short, and the number of tokens actually increases after constructing the memory summaries (from 268.4 to 592.9 on average). For medium-length videos, the number of tokens decreases from 2048.2 to 1290.0, corresponding to a 37\% reduction. For long videos (30–60 minutes), the token count drops substantially from 9717.7 to 2045.6, achieving a 78.9\% reduction. These results demonstrate that longer videos yield higher token reduction rates, indicating that the proposed method becomes increasingly effective as video length grows.

\begin{table}[t]
\caption{Performance comparison under audio-unavailable and audio-available scenarios on the VideoMME benchmark.}
\centering
\renewcommand{\tabcolsep}{2mm}
\begin{tabular}{ccc}
\toprule
\makecell{\textbf{Memory Type}} & 
\makecell{\textbf{Audio} \\ \textbf{Used}} & 
\makecell{\textbf{Performance} \\ \textbf{(\%)}} \\
\midrule
None (Baseline) & \xmark & 49.9 \\
Visual Caption-based & \xmark & 63.7 \\
Audio Transcript-based & \cmark & 73.4 \\
\bottomrule
\end{tabular}
\label{tab:3}
\end{table}

\subsubsection{Performance Evaluation under Different Memory Construction Scenarios}
To comprehensively evaluate the framework’s performance under different episodic memory construction scenarios, we compare two settings: one where episodic memory is derived from speech transcripts, and another where it is constructed from visual captions in the absence of audio inputs. In the latter case, we sample one frame every 10 seconds from each video and generate visual captions using the Qwen2.5-VL 7B model. (These captions are then used to build episodic memory, from which question-related information is retrieved to assess the framework’s robustness under the audio-unavailable condition.)

For this, we use a baseline model without episodic memory to measure the performance gain achieved by integrating different memory construction scenarios. As shown in Table~\ref{tab:3}, the visual caption-based memory substantially outperforms this baseline (63.7\% vs. 49.9\%), confirming that the proposed schematic and narrative-aware memory effectively compensates for the absence of audio inputs. Furthermore, the speech transcripts-based memory achieves a higher accuracy (73.4\%), illustrating the additional contribution of auditory cues. These results demonstrate the robustness of the proposed framework and the versatility of its memory construction scenarios.

\begin{table}[t]
\caption{Token compression between full transcripts and episodic memory on the Video-MME benchmark.}
\centering
\renewcommand{\tabcolsep}{2mm}
\begin{tabular}{cccc}
\toprule
\makecell{\textbf{Video Duration} \\ \textbf{(min)}} & 
\makecell{\textbf{\# Transcripts} \\ \textbf{Tokens}} & 
\makecell{\textbf{\# Memory} \\ \textbf{Tokens}} & 
\makecell{\textbf{Token Reduction} \\ \textbf{(\%)}} \\
\midrule
0--2 & 268.4 & 592.9 & --120.9 \\
4--15 & 2048.2 & 1290.0 & 37.0 \\
30--60 & 9717.7 & 2045.6 & 78.9 \\
\bottomrule
\end{tabular}
\label{tab:5}
\end{table}

\section{Conclusion}
In this work, we proposed GCAgent, the global-context-aware agent framework for long-video understanding that constructs episodic memory before queries. Departing from prior long-video MLLM pipelines that either extend model context or solely rely on on-demand retrieval, GCAgent explicitly builds schematic and narrative structures, mirroring human event cognition. It also operationalizes them within a perception–action–reflection paradigm. This design bridges the long-standing gap between global context modeling and query-conditioned interaction: the Memory Manager Agent forms structured episodic memory from transcripts, while the Reasoning Agent grounds inference on both the retrieved multimodal evidence and the global narrative context. Extensive experiments demonstrate that this global–local synergy consistently boosts reasoning accuracy and remains increasingly beneficial as video duration grows. We believe that grounding computational reasoning in cognitively-motivated event structure opens a more general path toward human-like long-video understanding beyond simple token scaling or agent orchestration alone.

\section{Discussion}
Our framework has demonstrated strong effectiveness in long video understanding; however, it still has several limitations. (i) The performance gain is less pronounced in the Multilingual domain. This limitation likely arises from semantic distortions introduced during the process of converting multilingual subtitles into episodic memory, where translation into English is performed. (ii) The current framework constructs episodic memory primarily based on either speech transcripts or visual captions, rather than integrating both modalities simultaneously. Extending the framework toward a multimodal memory construction approach could further enhance narrative coherence and overall understanding of long-form video content. (iii) Computational overhead is dominated by the memory manager agent. Although effective, it remains costly for extremely long videos. Developing a lightweight memory manager could lower the overall cost while maintaining memory quality.


\bibliographystyle{IEEEtran}
\bibliography{main}







\vfill

\clearpage

\begin{figure*}[t]
\centering
\centerline{\includegraphics[width=16cm]{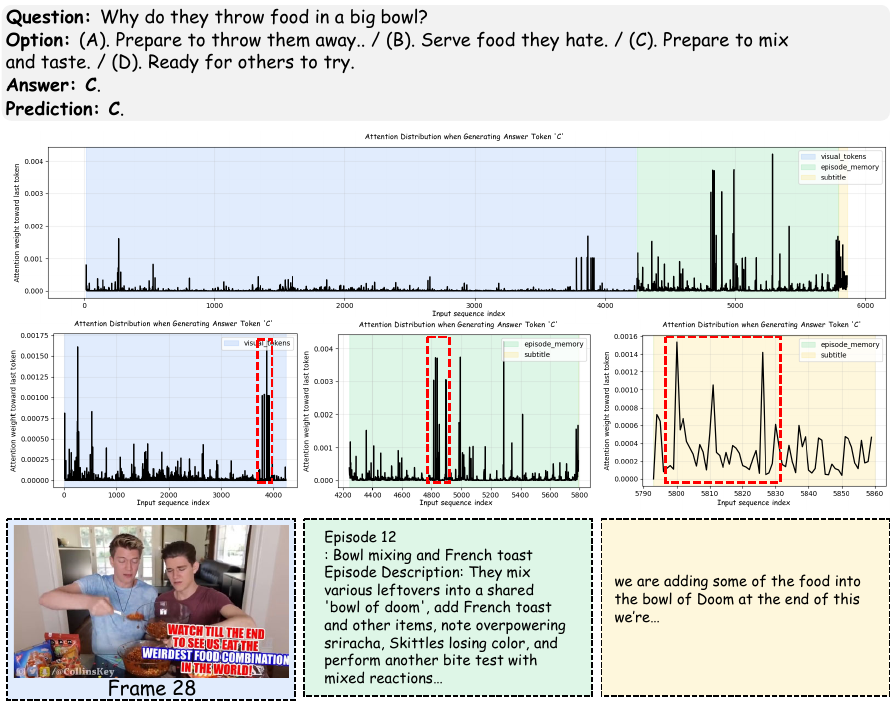}}
\caption{Visualization of attention distribution when predicting the answer for the question “Why do they throw food in a big bowl?”. The model attends to relevant visual frames, episodic memory (Episode 12), and query-related transcripts describing food mixing.}
\label{fig:4}
\end{figure*}

\section{Additional Results}
\subsection{Qualitative Result}
To qualitatively assess our framework’s capability in long video understanding, we analyze the average attention scores from the final layer across all heads with respect to the LLM input sequence index. Specifically, we examine the attention distribution of the reasoning agent at the time it predicts the answer token, measuring how much it attends to the query-related video segment, episodic memory, and query-related speech transcripts that are informative for answering the question.

As illustrated in Fig.~\ref{fig:4}, the question “Why do they throw food in a big bowl?” and the model’s correct prediction of the ground-truth answer “(C) Prepare to mix and taste” are shown. Below it, we visualize the average attention scores across the entire input sequence index. To further examine which sources the model attends to within each modality, we additionally provide three separate graphs corresponding to visual tokens, episodic memory, and textual context.

For the visual tokens extracted from query-related video segment, attention peaks at frame 28. This frame shows participants mixing food in a large bowl, with an on-screen caption “watch till the end to see us eat the weirdest food combination in the world,” which directly supports the reasoning behind (C).

For episodic memory, the model focuses on Episode~12 (“Bowl mixing and French toast”). This episode describes similar actions: mixing various leftovers into a shared “bowl of doom,” adding French toast and other ingredients, commenting on the taste of sriracha and the fading color of Skittles, and performing another bite test with mixed reactions. This is semantically aligned with the behavioral context implied by the question. 

For the text tokens extracted from query-related speech transcripts, they also receive high attention weights around phrases such as “we are adding some of the food into the bowl of doom at the end of this we’re…,” which occurs toward the end of the dialogue. This shows that the model grounds its reasoning in linguistic cues that directly describe the final action of adding food into the shared bowl. Overall, the model consistently attends to semantically coherent and question-relevant information across modalities when reasoning over long videos.

\begin{figure*}[t]
\centering
\centerline{\includegraphics[width=17cm]{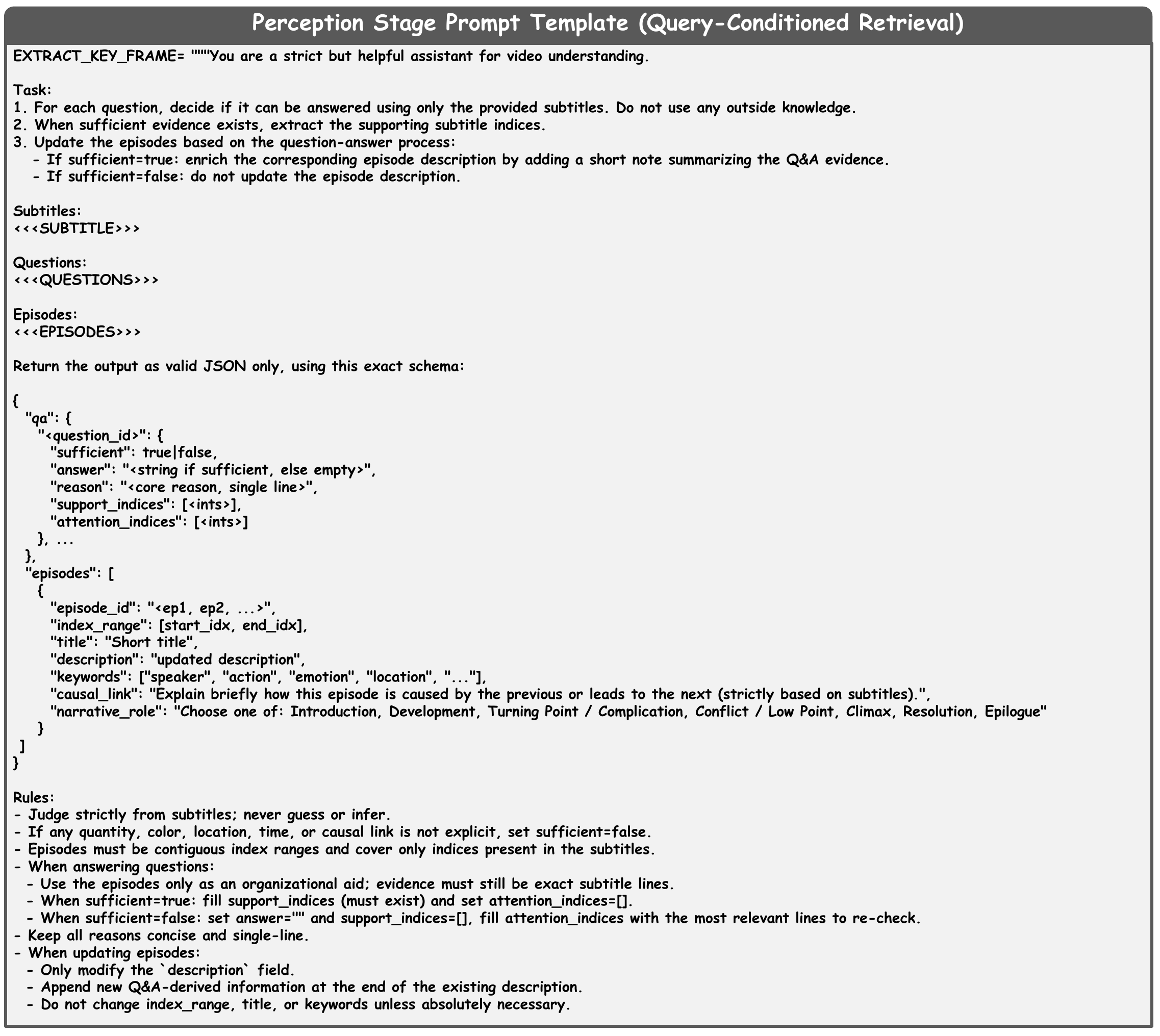}}
\caption{Perception prompt template used for query-conditioned retrieval.}
\label{fig:5}
\end{figure*}

\begin{figure*}[t]
\centering
\centerline{\includegraphics[width=17cm]{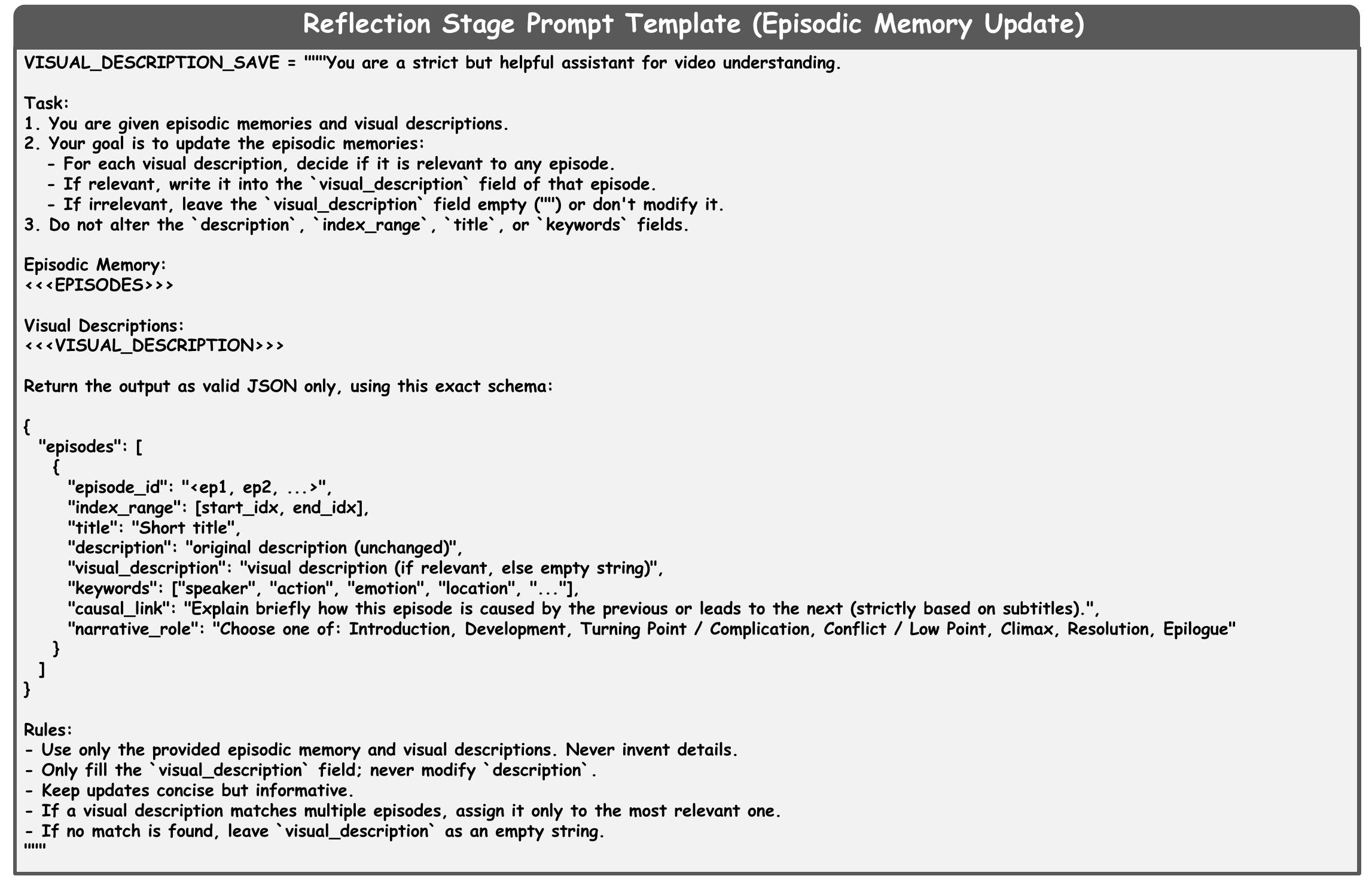}}
\caption{Reflection prompt template used for episodic memory update.}
\label{fig:6}
\end{figure*}

\section{Prompt Templates for Perception and Reflection}
Fig.~\ref{fig:5} and Fig.~\ref{fig:6} present the exact prompt templates used by the memory manager agent in the Perception and Reflection stages, respectively. These prompts operationalize the two stages described in method section: the perception prompt reads the query and locates the minimally sufficient transcript spans (with temporal indices) required to answer the query, while the reflection prompt takes the predicted answer and its retrieved visual/textual evidence and compresses them into an event-level narrative summary that is appended back to episodic memory. We disclose these verbatim prompts below to ensure full reproducibility.

\end{document}